\documentclass{article}
\usepackage{enumitem}
\usepackage{spconf,amsmath,graphicx}
\usepackage{xcolor}
\usepackage{caption}
\usepackage{subcaption}
\usepackage{hyperref}

\title{Sequence Model with Self-Adaptive Sliding Window for Efficient Spoken Document Segmentation}
\name{Qinglin Zhang$^1$, Qian Chen$^1$, Yali Li$^1$, Jiaqing Liu$^2$, Wen Wang$^1$}
\address{$^1$Speech Lab, Alibaba Group. $^2$Renmin University of China.\\
  \small\texttt{\{qinglin.zql, tanqing.cq, chuzhou.lyl, w.wang\}@alibaba-inc.com, jiaqingliu@ruc.edu.cn}}
\begin{document}
%
\maketitle
\begin{abstract}
Transcripts generated by automatic speech recognition (ASR) systems for spoken documents lack structural annotations such as paragraphs, significantly reducing their readability.  Automatically predicting paragraph segmentation for spoken documents may both improve readability and downstream NLP performance such as summarization and machine reading comprehension. We propose a sequence model with self-adaptive sliding window for accurate and efficient paragraph segmentation. We also propose an approach to exploit phonetic information, which significantly improves robustness of spoken document segmentation to ASR errors.  Evaluations are conducted on the English Wiki-727K document segmentation benchmark, a Chinese Wikipedia-based document segmentation dataset we created, and an in-house Chinese spoken document dataset. Our proposed model outperforms the state-of-the-art (SOTA) model based on the same BERT-Base, increasing segmentation F1 on the English benchmark by 4.2 points and on Chinese datasets by 4.3-10.1 points, while reducing inference time to less than 1/6 of inference time of the current SOTA.
\end{abstract}

\section{Introduction}
\label{sec:introduction}
The number of spoken documents has been constantly increasing in the forms of recorded meetings, lectures, interviews, etc. The transcripts generated by automatic speech recognition (ASR) systems for spoken documents lack structural annotations such as paragraphs. Paragraphs or sections are semantically coherent sub-document units, usually corresponding to topic or sub-topic boundaries~\cite{10.1145/584792.584829}.  In addition, downstream natural language processing (NLP) applications such as summarization and machine reading comprehension are typically trained on well-formed text with paragraph segmentation. Lack of paragraph segmentation in ASR transcripts significantly reduces readability of the transcripts and may also dramatically degrade the performance of downstream applications due to mismatched data conditions ~\cite{DBLP:conf/hicss/BoguraevN00}. 

\emph{Document segmentation} is defined as automatic prediction of segment (paragraph or section) boundaries for a document~\cite{hearst-1994-multi}. Previous document segmentation works mostly focused on written text and include both unsupervised~\cite{DBLP:conf/anlp/Choi00,DBLP:conf/naacl/Eisenstein09,glavas-etal-2016-unsupervised} and supervised approaches~\cite{koshorek-etal-2018-text,DBLP:conf/ijcai/LiSJ18, DBLP:journals/corr/abs-1808-09935}. Recently many neural approaches were proposed for document segmentation. \cite{DBLP:conf/emnlp/LukasikDPS20} proposed a cross-segment BERT model using local context and a hierarchical BERT model (Hier.BERT) using two BERT models to encode sentence and document separately for exploiting longer context. \cite{DBLP:conf/aaai/GlavasS20} also uses two hierarchically connected transformers. However, hierarchical models are computationally expensive and slow for inference. We aim at finding a good balance between exploring sufficient context for accurate segmentation while attaining high inference efficiency. Our contributions can be summarized as follows.
\begin{itemize}[leftmargin=*,noitemsep]
\item We propose a sequence model (denoted by \textbf{SeqModel}) for modeling document segmentation as a sentence-level sequence labeling task. Our SeqModel encodes longer context without hierarchical encoding. We observe that SeqModel using pre-trained models with enhanced structural modeling improves segmentation accuracy. 
\item  We propose a self-adaptive sliding window approach that further improves inference efficiency.
\item We propose a phone-embedding based approach and improve spoken document segmentation F$_1$ by 2.1-2.8 points.
\item Systematic evaluations on the English Wiki-727K benchmark and a Chinese Wikipedia dataset (Wiki-zh)\footnote{\url{https://drive.google.com/file/d/1o1kQrCUMIg3WZSqfMiEybEJwkWf\_oHtC}
} created in this work demonstrate that SeqModel based on BERT-Base significantly outperforms the current state-of-the-art (SOTA) model cross-segment BERT-Base~\cite{DBLP:conf/emnlp/LukasikDPS20} with the same number of model parameters, improving segmentation F$_1$ by 4.2 points on the Wiki-727K test set, by 4.3 points on the Wiki-zh test set, and by 8.8-10.1 points on a Chinese spoken document dataset. SeqModel achieves 81.7\% relative inference speedup over cross-segment BERT-Base; adding the proposed self-adaptive sliding window yields additional 9.6\% relative speedup and reduces the inference time of SeqModel to less than 1/6 of the inference time of cross-segment BERT-Base. 
\end{itemize}

\begin{figure*}[!t]
\centering
\begin{subfigure}[b]{0.7\textwidth}
\centering
\includegraphics[width=\textwidth]{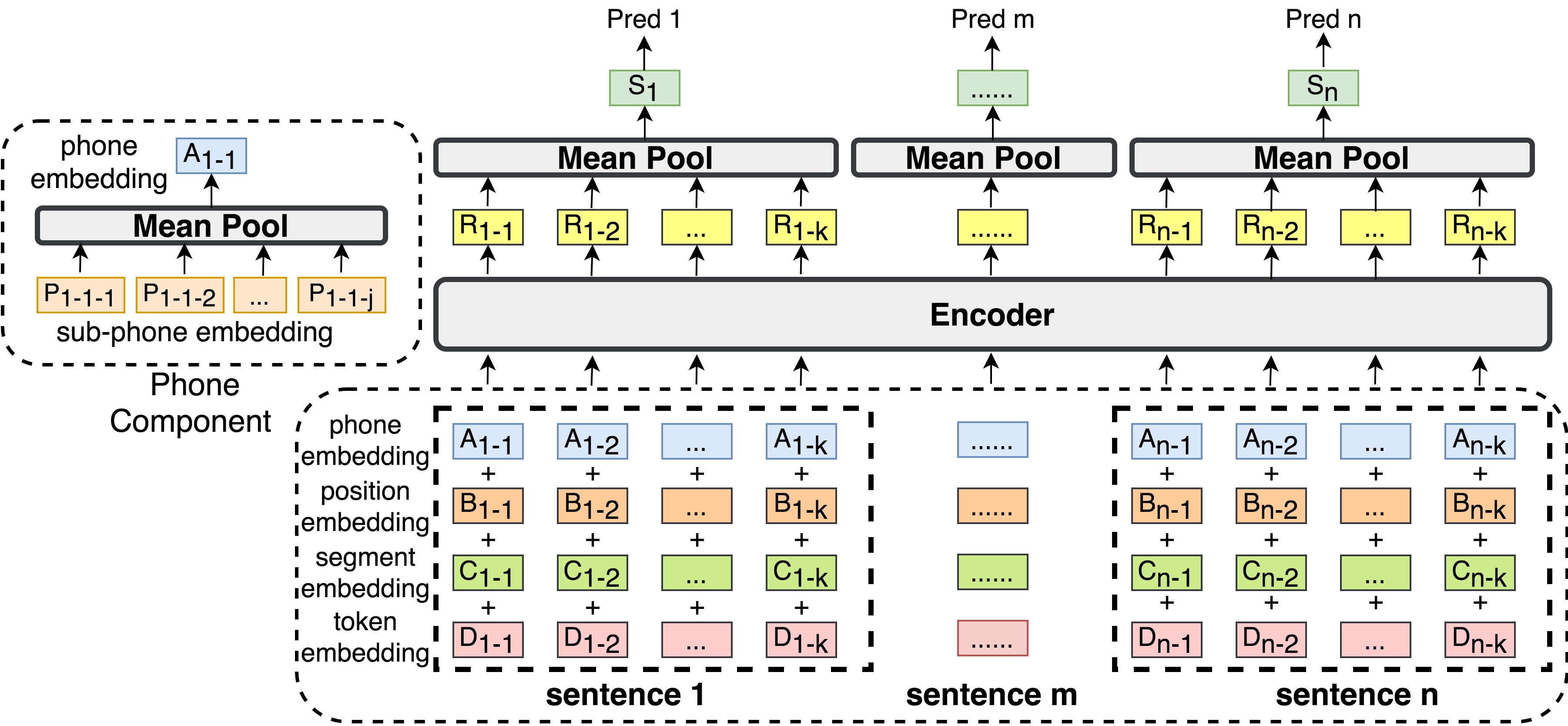}
  \caption{A diagram illustrating our proposed \textbf{SeqModel} for document segmentation. The phone component on the left provides an enlarged view for illustrating how phone embeddings are computed to improve the robustness of the proposed model for segmenting spoken documents (Section~\ref{subsec:phoneembedding}).}
  \label{fig:modelarchitecture}
\end{subfigure}
\begin{subfigure}[b]{0.7\textwidth}
  \centering
  \includegraphics[width=\textwidth]{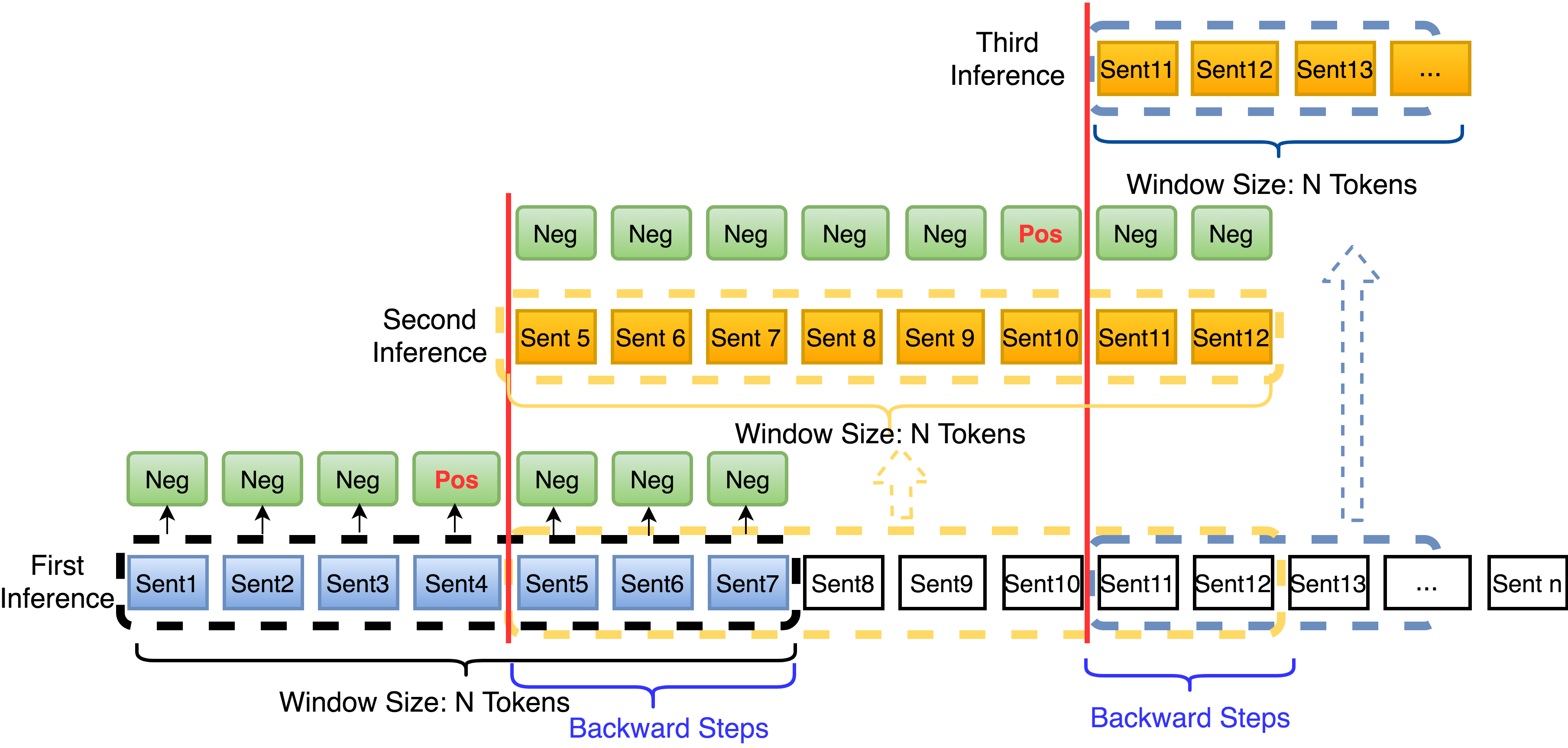}
  \caption{The proposed self-adaptive sliding window for further speeding up inference for SeqModel.}
  \label{fig:sliding-window}
\end{subfigure}
\caption{The proposed SeqModel architecture and the self-adaptive sliding window approach for inference.}
\label{fig:all}
\end{figure*}

\section{Proposed Model}
\label{sec:model}

\subsection{Sequence Model}
\label{subsec:sequencemodel}
Figure~\ref{fig:modelarchitecture} illustrates the proposed \textbf{SeqModel}, which models document segmentation as a sentence-level sequence labeling task. Each sentence in a document is first tokenized by WordPiece tokenizer~\cite{DBLP:journals/corr/WuSCLNMKCGMKSJL16}. A block of consecutive sentences $\{s_m\}_{m=1}^{n}$ is prepended with a special token [CLS]. This sequence is embedded through the input representation layer as the element-wise sum of token embedding, position embedding, and segment embedding. These embeddings are then fed to a transformer encoder and the output hidden states corresponding to $k$ tokens in sentence $s_m$, denoted by $\{R_{m-i}\}_{i=1}^{k}$,  are mean-pooled to represent the sentence encoding. The sentence encodings $\{S_m\}_{m=1}^{n}$ for sentences $\{s\}_{m=1}^{n}$ are fed to a softmax binary classifier to classify whether each sentence is a segment boundary. The training objective is minimizing the softmax cross-entropy loss.

One SOTA model, the cross-segment BERT model~\cite{DBLP:conf/emnlp/LukasikDPS20},  treats document segmentation as a token-level binary classification task for each candidate break (i.e., each token). The input training sample is composed of a leading [CLS] token, and concatenation of the left and right local contexts for a candidate break, separated by the [SEP] token. This sequence is fed to a BERT-encoder and the output hidden state corresponding to [CLS] is fed to a softmax classifier for segmentation decision on the candidate break. In comparison, our SeqModel encodes a block of sentences $\{s_m\}_{m=1}^{n}$ simultaneously, hence exploits longer context and inter-sentence dependencies through encoder self-attention. Consequently, the contextualized sentence encodings  $\{S_m\}_{m=1}^{n}$ for classification can potentially improve segmentation accuracy. In addition, our SeqModel  classifies multiple sentences simultaneously, hence significantly speeds up inference compared to the cross-segment model. Also, different from another SOTA model Hier.BERT~\cite{DBLP:conf/emnlp/LukasikDPS20}, SeqModel generates contextualized sentence encoding through mean-pooling, instead of employing another transformer encoder for document encoding as in Hier.BERT, hence significantly reduces computational complexity and improves inference efficiency.

Document segmentation requires deep understanding of document structure. We hypothesize that using pre-trained models with enhanced structural modeling for SeqModel may improve segmentation accuracy. 
BERT~\cite{DBLP:conf/naacl/DevlinCLT19} employs the next sentence prediction (NSP) task as an inter-sentence objective, which is a binary classification task deciding whether two sentences are contiguous in the original source. RoBERTa~\cite{DBLP:journals/corr/abs-1907-11692} removes the NSP objective from pre-training but also adds other training optimizations. StructBERT~\cite{DBLP:conf/iclr/0225BYWXBPS20} augments BERT pre-training objectives with a word structural objective and a sentence structural objective. 
The sentence structural objective conducts a ternary classification on two sentences $(S_1,S_2)$ to decide whether $S_1$ precedes or follows $S_2$ or the two sentences are noncontiguous. The CONPONO model~\cite{DBLP:conf/acl/IterGLJ20} exploits new inter-sentence objectives, including predicting ordering between coherent yet noncontiguous spans with multiple sentences in between. ELECTRA~\cite{DBLP:conf/iclr/ClarkLLM20} replaces masked language modeling in BERT pre-training with a more efficient replaced-token detection task, which replaces some input tokens with plausible alternatives generated by a generator network, then trains a discriminator to predict whether each token in the corrupted input is replaced by a generated sample or not. We include RoBERTa and ELECTRA in investigation due to their strong performance, but they do not enhance structural modeling particularly.

\vspace{-2mm}
\subsection{Self-adaptive Sliding Window}
\label{subsec:inference}
We also propose a self-adaptive sliding window approach to further speed up inference without sacrificing accuracy, as depicted in Figure~\ref{fig:sliding-window}. Traditional sliding window for segmentation inference uses a fixed forward step size. 
In our proposed self-adaptive sliding window approach, during inference, starting from the last sentence in the previous window, the model looks backward within a maximum backward step size to find positive segmentation decisions (segmentation prediction probability from the model $>0.5$) from the previous inference step. When there are positive decisions within this span, the next sliding window is automatically adjusted to start from the next sentence after the most recent predicted segment boundary. Otherwise, when no positive decisions exist within this span, the next sliding window starts from the last sentence in the previous window. Considering that the last paragraph and the history beyond have reduced impact on the next segmentation decision, this strategy helps discard irrelevant history within the sliding window. Hence, the self-adaptive sliding window may both speed up inference and improve segmentation accuracy.

For training, we compare two approaches: constructing samples following the self-adaptive sliding window approach, but using the segmentation references instead of predicted positive segmentation decisions from the model; or constructing samples with fixed sliding window. Compared to using the self-adaptive sliding window with segmentation references, the fixed sliding window approach is more effective at reducing the discrepancy between training and inference, creates more training samples, and ultimately ameliorates robustness of the segmentation model during inference to noises from wrong segment boundary predictions from the model. We observe training with fixed sliding window significantly outperforms training with self-adaptive sliding window using segmentation references, on segmentation F$_1$. 
In order to make SeqModel learn inter-sentence dependencies among as many sentences as possible and become robust to sentences with missing tokens,  
for training sample sequences longer than the max sequence length, 
we first truncate them beyond the max sequence length. We also create an alternative by truncating tokens in sentences of the sequence while keeping the max sentence number in the sequence. We pool these two versions for training. For training SeqModel, we optimize forward step size for fixed sliding window and the max sentence number, based on segmentation F$_1$ on Wiki-727K dev.

\vspace{-3.5mm}
\subsection{Phone Embedding}
\label{subsec:phoneembedding}
Inputs to SeqModel for spoken document segmentation are ASR 1-bests. ASR errors usually comprise acoustically confusing words with similar pronunciations yet different meanings, which become noise to the input embedding layer of SeqModel. We propose an approach to improve robustness to ASR errors by augmenting the input embeddings with phone embeddings (denoted by \emph{PEs}), as shown in the input embedding layer in Figure~\ref{fig:modelarchitecture}. PEs also serve a regularization to the original input embedding. Details for computing PEs are illustrated in the enlarged view of the phone component in Figure~\ref{fig:modelarchitecture}\footnote{We compared computing the PEs as the sum or the mean of sub-phone embeddings and observed consistently better segmentation accuracy from mean-pooling.}.
Given a word (use the first word $w_1$ in $s_1$ as an example) and its phone sequence $\{p_{1-1-k}\}_{k=1}^{j}$ (from looking up an in-house Chinese pronunciation dictionary and picking up the first entry for the word), the sub-phone embeddings are denoted by $\{P_{1-1-k}\}_{k=1}^{j}$. The augmented input embedding $E[w_1,\{p_{1-1-k}\}_{k=1}^{j}]$ for $w_1$ is computed as $E[w_1,\{p_{1-1-k}\}_{k=1}^{j}] = E[w_1] + \frac{1}{j}\sum_{k=1}^{j}E[p_{1-1-k}]$,
where $E(*)$ denotes trainable embedding; $E[w_1]$ denotes the standard 
BERT-style input embedding. In Figure~\ref{fig:modelarchitecture}, $\{P_{1-1-k}\}_{k=1}^{j}$ denotes sub-phone embeddings $\{E[p_{1-1-k}]\}_{k=1}^{j}$; $A_{1-1}$ denotes $\frac{1}{j}\sum_{k=1}^{j}E[p_{1-1-k}]$.

\vspace{-2mm}
\section{Experiments}
\label{sec:experiments}
\vspace{-1mm}
\subsection{Experimental Setup}
\label{subsec:datasets}
We conduct document segmentation evaluations on three datasets, including the commonly used English Wiki-727K benchmark, a Chinese Wikipedia-based document segmentation dataset that we created in this work (denoted by \emph{Wiki-zh}), and an in-house Chinese spoken document dataset. 
\newline
\noindent\textbf{Wiki-727K and Wiki-zh.} The Wiki-727K dataset~\cite{koshorek-etal-2018-text} comprises 727K English Wikipedia articles. We use the same train\slash dev\slash test partitions as in the original work~\cite{koshorek-etal-2018-text}. Note that English Wiki-727K uses the hierarchical segmentation of articles as in their table of contents, that is, the segment boundaries are section boundaries.  Due to lack of Chinese document segmentation benchmarks, we create the Wiki-zh dataset in a similar procedure as used for creating Wiki-727K except one difference. Given that Chinese wikipedia articles are much shorter than English wikipedia articles\footnote{\url{https://linguatools.org/tools/corpora/wikipedia-monolingual-corpora/}}, for Wiki-zh, to make the average number of segments per document comparable to that of Wiki-727K, we define segment boundaries as paragraph boundaries. From a Chinese Wikipedia snapshot, we extract and tokenize documents and define reference segment boundaries based on double new lines.
\newline 
\noindent\textbf{Building the Chinese spoken document dataset.} Public spoken document segmentation datasets are particularly scarce. Our in-house Chinese spoken document dataset consists of two subsets as single-speaker lectures (denoted \emph{SD-zh-SP}) and multi-speaker interviews (denoted \emph{SD-zh-MP}), all from the finance domain. We use an in-house open-domain ASR system (CER 16.0\% on in-house testsets) to generate ASR 1-best for SD-zh-SP and SD-zh-MP. We employ transformer-based punctuation prediction to insert period, question mark, and comma into the ASR 1-bests. The resulting transcripts are used for manual annotations through crowd-sourcing, by annotating whether each sentence (based on the automatically inserted period or question mark) is at the paragraph boundary. However, these ASR 1-bests contain ASR errors, ungrammatical 
structures, disfluencies
, etc. All these issues pose severe challenges to manual annotations. We observe significant inter-annotator disagreements on paragraph annotations. Initially, only 5\% annotated sentences have the same labels from multiple annotators. 

We tackle this challenge with a three-stage approach, namely, screening, annotation, and post-annotation filtering.  During screening, the annotators are asked to annotate paragraphs for formal news text (with reference paragraph boundaries removed). Annotations from the annotators are scored against the reference paragraph boundaries. Only annotators with F$_1 > 60$ are selected for annotating the spoken documents. During annotation,  we collect annotations from 5 annotators for each sentence.  During post-annotation filtering,  we use leave-one-out to score the annotations from each annotator. The top4 annotations for each sentence are used for majority voting to make the final segmentation decision for each sentence ($\ge 3$ positive votes count a positive decision). We observe significant improvement on annotation quality from this approach. On formal news text, the average annotation F$_1$ improves from 61.8 to 68. On the combined set of SD-zh-SP and SD-zh-MP, the average F$_1$ from majority voting on 5 annotations is 33.4. Using the top4 annotations after leave-one-out improves F$_1$ to 39.5. Still, the gap between the average F$_1$ on written text and spoken documents indicates the challenges for spoken document segmentation annotations, which demands more studies.

\begin{table}[htb]
\renewcommand{\arraystretch}{0.9}
\begin{center}
\scalebox{0.9}{
\begin{tabular}{l c c c c c}
\hline
\textbf{Dataset} & \textbf{Docs} & \textbf{\#S\slash doc} & \textbf{SLen} & \textbf{\#P\slash doc} & \textbf{\#S\slash P}\\
\hline
Wiki-727K train     & 582,160 & 47.9 & 20.7 	& 5.3 	& 9.1 \\
Wiki-727K dev       & 72,354	& 48.1 	& 20.8 	& 5.3 	& 9.1 \\
Wiki-727K test      & 73,232	& 48.7 	& 20.7 	& 5.3 	& 9.1 \\
\hline
\hline
Wiki-zh  train	    & 656,602	& 11.9 & 38.8 	& 4.7 	& 2.5 \\
Wiki-zh  dev	    & 82,011	    & 11.8 & 38.7 	& 4.7 	& 2.5 \\
Wiki-zh  test	    & 82,160	    & 11.9 & 38.8 	& 4.7 	& 2.5  \\
\hline
\hline 
SD-zh-SP            & 94	    & 209.3 & 34.9 	& 12.4 	& 16.9 \\
SD-zh-MP            & 4974	    & 8.1 	& 27.4 	& 1.2 	& 6.7 \\        
\hline
\end{tabular}
}
\end{center}
\vspace{-1mm}
\caption{Data statistics. For each dataset, \textbf{Docs} denotes the total number of documents; \textbf{\#S\slash doc} and \textbf{\#P\slash doc}  denote the average number of sentences and paragraphs per document, respectively; \textbf{SLen} denotes the mean sentence length; \textbf{\#S\slash P} denotes the average number of sentences per paragraph.}
\label{tab:statistics}
\end{table}

\begin{table}[htb]
\renewcommand{\arraystretch}{0.9}
\begin{center}
\begin{tabular}{l c c c}
\hline
\textbf{Model} & \multicolumn{3}{c}{\textbf{Wiki-727K}} \\
& \textbf{P} & \textbf{R} & \textbf{F$_1$} \\
\hline
BERT-Base + Bi-LSTM                 & 67.3 & 53.9 & 59.9 \\
Hier. BERT (24-Layers)              & 69.8 & 63.5 & 66.5 \\
Cross-segment BERT-Base 128-128     & - & - & 64.0 \\
Cross-segment BERT-Large 128-128   & 69.1 & 63.2 & 66.0 \\
Cross-segment BERT-Large 256-256    & 61.5 & 73.9 & 67.1 \\
\hline
\hline
Our SeqModel:BERT-Base                
&70.6 & 65.9 & \textbf{68.2} \\
\hline
\end{tabular}
\end{center}
\vspace{-1mm}
\caption{Precision, recall, and F$_1$ on the English document segmentation benchmark Wiki-727K test set from prior models and our proposed SeqModel based on BERT-Base. The first group of results in the table are all cited from~\cite{DBLP:conf/emnlp/LukasikDPS20}.}
\label{tab:compare-sota-en}
\end{table}

\begin{table*}[htb]
\renewcommand{\arraystretch}{0.9}
\begin{center}
\begin{tabular}{l c c c | c c c | c c c}
\hline
\textbf{Model}
& \multicolumn{3}{c|}{\textbf{Wiki-zh}}
& \multicolumn{3}{c|}{\textbf{SD-zh-SP}}
& \multicolumn{3}{c}{\textbf{SD-zh-MP}} \\
& \textbf{P} & \textbf{R} & \textbf{F$_1$}
& \textbf{P} & \textbf{R} & \textbf{F$_1$}
& \textbf{P} & \textbf{R} & \textbf{F$_1$} \\
\hline

Cross-segment BERT-Base 128-128
& 61.2 	& 80.2 & 69.4
& 14.8 	& 35.7  & 21.0
& 8.2 	& 28.1 & 13.2 \\
\hline
\hline
SeqModel:BERT-Base
& 78.4 	& 69.5 & \textbf{73.7}
& 28.4 	& 31.4 & \textbf{29.8} 
& 18.3 	& 32.1 & \textbf{23.3} \\
\hline
\end{tabular}
\end{center}
\caption{Precision, recall, and F$_1$ on the Chinese Wiki-zh test set and SD-zh-SP and SD-zh-MP datasets from Cross-segment BERT-Base 128-128 and our proposed SeqModel based on BERT-Base.}
\label{tab:compare-sota-zh}
\end{table*}

\begin{table*}[htb]
\renewcommand{\arraystretch}{0.9}
\begin{center}
\begin{tabular}{l c c c | c c c | c c c | c c c}
\hline
\textbf{Model} & \multicolumn{3}{c|}{\textbf{Wiki-727K}}
& \multicolumn{3}{c|}{\textbf{Wiki-zh}}
& \multicolumn{3}{c|}{\textbf{SD-zh-SP}}
& \multicolumn{3}{c}{\textbf{SD-zh-MP}} \\
& \textbf{P} & \textbf{R} & \textbf{F$_1$} 
& \textbf{P} & \textbf{R} & \textbf{F$_1$}
& \textbf{P} & \textbf{R} & \textbf{F$_1$}
& \textbf{P} & \textbf{R} & \textbf{F$_1$} \\
\hline
SeqModel:BERT-Base                
&70.6 & 65.9 & 68.2
& 78.4 	& 69.5 & 73.7
& 28.4 	& 31.4 & 29.8 
& 18.3 	& 32.1 & 23.3 \\
SeqModel:StructBERT-Base                 & 71.9 & 68.9 & 70.4
& 79.2 	& 72.7 & \textbf{75.8}
& 32.9 & 43.0 & \textbf{37.3} 
& 19.2 & 30.9 & 24.6 \\
SeqModel:CONPONO-Base                & 71.1 & 65.9 & 68.4
& - & - & -
& - & - & -
& - & - & - \\
SeqModel:RoBERTa-Base                & 66.2 & 74.7 & 70.2
& 74.6 	& 73.7 & 74.2
& 25.6 & 45.1 & 32.7
& 16.7 & 44.9 & 24.3 \\
SeqModel:ELECTRA-Base               & 72.3 & 71.1 & \textbf{71.7}
& 73.5 	& 76.6 & 75.0
& 31.1 & 44.0 & 36.4
& 20.6 & 37.4 & \textbf{26.6} \\
\hline
\end{tabular}
\end{center}
\caption{Precision\slash recall\slash F$_1$ on the Wiki-727K and Wiki-zh test sets and SD-zh-SP and SD-zh-MP datasets from our proposed SeqModel based on different pre-trained model MODEL, denoted by ``SeqModel:MODEL''. Note that since a CONPONO pre-trained model for Chinese is not available, no results are reported for SeqModel:CONPONO-Base on all Chinese datasets.}
\label{tab:compare-pretrainedmodels}
\end{table*}

\begin{table*}
\renewcommand{\arraystretch}{0.9}
\begin{center}
\begin{tabular}{l c c c | c c c | c c c}
\hline
\textbf{Model}
& \multicolumn{3}{c|}{\textbf{Wiki-zh}}
& \multicolumn{3}{c|}{\textbf{SD-zh-SP}}
& \multicolumn{3}{c}{\textbf{SD-zh-MP}} \\
& \textbf{P} & \textbf{R} & \textbf{F$_1$}
& \textbf{P} & \textbf{R} & \textbf{F$_1$}
& \textbf{P} & \textbf{R} & \textbf{F$_1$} \\
\hline
(a) SeqModel:BERT-Base                
& 78.4 	& 69.5 & 73.7
& 28.4 	& 31.4 & 29.8 
& 18.3 	& 32.1 & 23.3 \\
(b) (a)+phone                       
& 78.9 & 70.2 & 74.3
& 29.4 & 36.6 & 32.6 
& 18.9 & 38.7 & 25.4 \\
(c) SeqModel:StructBERT-Base                 
& 79.2 	& 72.7 & 75.8
& 32.9 & 43.0 & 37.3 
& 19.2 & 30.9 & 24.6 \\
(d) (c)+phone                       
& 80.1 & 72.4 & 76.0
& 31.4 & 45.7 & 37.2
& 19.4 & 36.7 & 25.4  \\
\hline
\end{tabular}
\end{center}
\caption{Effect of adding phone embeddings (denoted by \textbf{+phone}) on segmentation accuracy on the Chinese spoken document SD-zh-SP and SD-zh-MP datasets and the written text Wiki-zh test set, evaluated on SeqModel:BERT-Base and SeqModel:StructBERT-Base. ``SeqModel:MODEL" denotes SeqModel based on the pre-trained model MODEL.}
\label{tab:phoneEmb}
\end{table*}
\vspace{-2mm}
\noindent\textbf{Data Statistics.} Table~\ref{tab:statistics} shows statistics of the three datasets. All Chinese datasets are tokenized with Jieba\footnote{\url{https://github.com/fxsjy/jieba}}. Due to the different segment granularity decision for Wiki-727K and Wiki-zh as explained earlier, \#P\slash doc for Wiki-727K and Wiki-zh are comparable but \#S\slash P are quite different, 9.1 versus 2.5. It is also noticeable that compared to SD-zh-MP, SD-zh-SP documents and human annotated paragraphs for SD-zh-SP are both much longer.
For Wiki-727K and Wiki-zh, we finetune the segmentation models on the train set, optimize hyperparameters on the dev set, and report segmentation accuracy on the test set, respectively. We use models trained on Wiki-zh train set with hyperparameters optimized on Wiki-zh dev for inference on SD-zh-SP and SD-zh-MP, due to the small size of the spoken document dataset.
\newline
\noindent\textbf{Evaluation Metrics.} As in~\cite{DBLP:conf/emnlp/LukasikDPS20}, we compute positive precision\slash recall\slash F$_1$, i.e., precision, recall, and F$_1$ based on the segment boundary references. To simplify notations, we use precision(P), recall(R), and F$_1$ throughout the paper to denote positive precision\slash recall\slash F$_1$. As in~\cite{DBLP:conf/emnlp/LukasikDPS20}, for all evaluations, we exclude the last boundary of each sentence\slash document as it is trivial to categorize them as segment boundaries.
\newline 
\noindent\textbf{Implementation Details.} The max sequence length is 512, batch size is 48, and SeqModel is trained for 2 epochs. Adam~\cite{DBLP:journals/corr/KingmaB14} is used for optimization. The initial learning rate is 5e-5 and the dropout probability is 0.1. For training, we choose fixed sliding window with forward step size 10 and max sentence number 60 by optimizing F$_1$ on Wiki-727K dev.

\vspace{-3mm}
\subsection{Results and Analyses}
\label{subsec:intrinsic}
\textbf{Comparison with previous SOTA.} Table~\ref{tab:compare-sota-en} shows P, R, and F$_1$ on the Wiki-727K test set comparing our SeqModel with prior models. The results for prior models, including BERT+Bi-LSTM, Hier.BERT and Cross-segment BERT-Base\slash BERT-Large with different left-right context lengths (128-128, 256-256), are all cited from \cite{DBLP:conf/emnlp/LukasikDPS20} and the best previous F$_1$ 67.1 on Wiki-727K is from Cross-segment BERT-Large 256-256. We reimplemented Cross-segment BERT-Base and replicated results on the Wiki-727 test set as in~\cite{DBLP:conf/emnlp/LukasikDPS20}. As can be seen from the table, the proposed SeqModel based on BERT-Base (denoted SeqModel:BERT-Base, 110M parameters) establishes a new SOTA F$_1$ 68.2, \textbf{4.2} points gain over Cross-segment BERT-Base 128-128 (F$_1$ 64.0) with the same number of parameters, and \textbf{1.1} points gain over Cross-segment BERT-Large 256-256 with much larger 336M model parameters. Table~\ref{tab:compare-sota-zh} compares the segmentation accuracy from SeqModel:BERT-Base and Cross-segment BERT-Base 128-128 on the Wiki-zh test set, SD-zh-SP, and SD-zh-MP. Again SeqModel:BERT-Base significantly outperforms Cross-segment BERT-Base 128-128 on the three Chinese test sets, by \textbf{4.3} (69.4 to 73.7), \textbf{8.8} (21.0 to 29.8), and \textbf{10.1} (13.2 to 23.3) points gains on F$_1$, respectively. All these improvements are statistically significant with $p < 0.0005$.

Table~\ref{tab:compare-pretrainedmodels} shows P, R, and F$_1$ from SeqModel using different pre-trained model MODEL, denoted by ``SeqModel:MODEL". Confirming our hypothesis, SeqModel:StructBERT-Base, based on StructBERT with enhanced modeling of inter-sentence relations, outperforms SeqModel:BERT-Base by +2.2 F$_1$ on Wiki-727K, +2.1 F$_1$ on Wiki-zh, +7.5 F$_1$ and +1.3 F$_1$ on SD-zh-SP and SD-zh-MP, respectively. There is only a minor +0.2 F$_1$ gain from SeqModel:CONPONO-Base over SeqModel:BERT-Base on Wiki-727K, although CONPONO-Base also extends inter-sentence objectives. Note that since a Chinese CONPONO pre-trained model is not available, no results are reported for SeqModel:CONPONO-Base on the three Chinese datasets. Although RoBERTa removes NSP from pre-training, SeqModel:RoBERTa-Base yields higher F$_1$ than SeqModel:BERT-Base on all testsets, due to other training optimizations. It is noticeable that although ELECTRA does not particularly enhance structural modeling, SeqModel:ELECTRA-Base attains the best F$_1$ on Wiki-727K and SD-zh-MP and second best F$_1$s on Wiki-zh and SD-zh-SP testsets, indicating that the more challenging and efficient pre-training task in ELECTRA is beneficial for SeqModel.

\begin{figure}
\centering
\begin{subfigure}{.25\textwidth}
  \centering
  \includegraphics[width=1.0\linewidth]{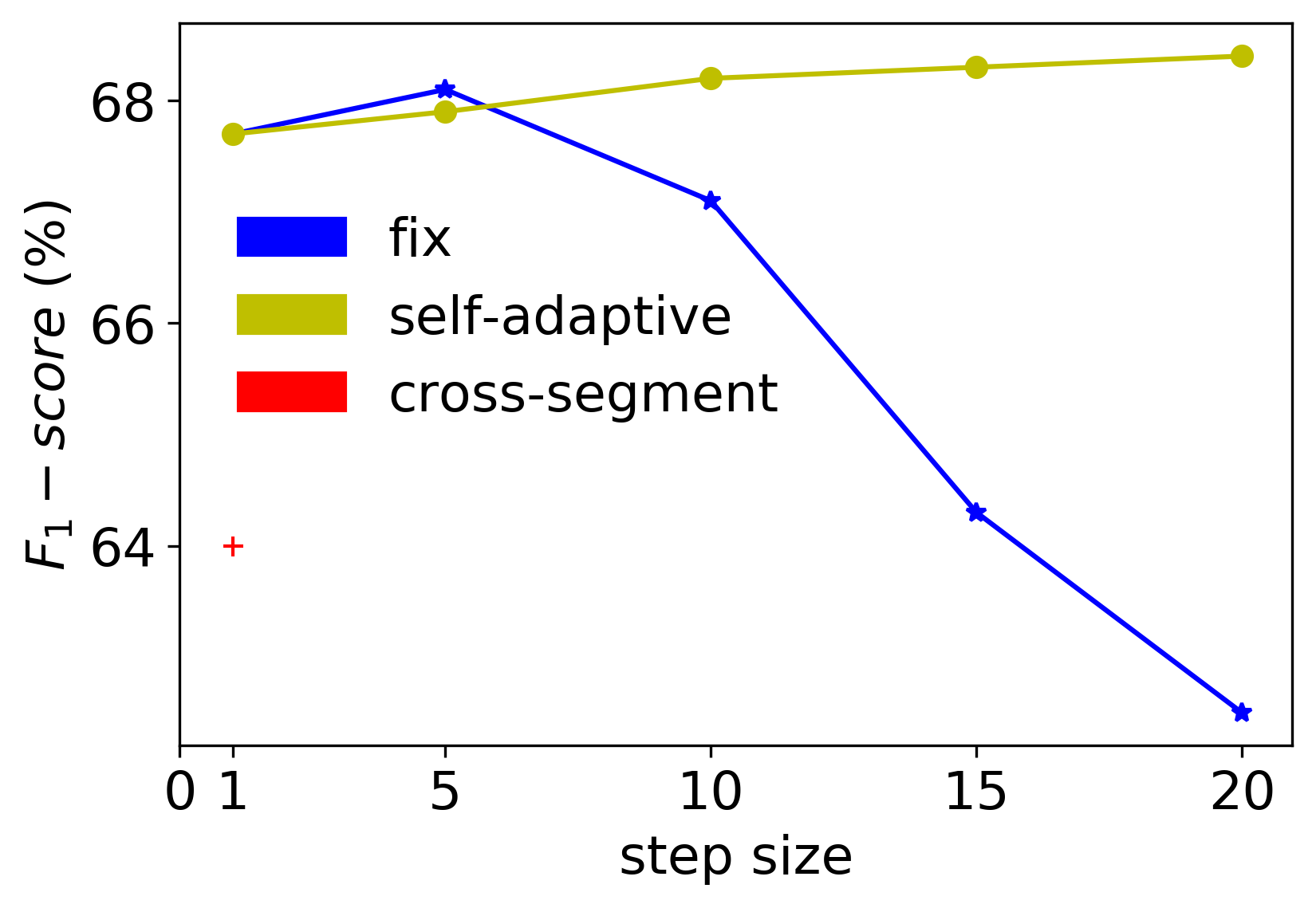}
  \caption{F$_1$ score VS step size.}
  \label{fig:f1score_windowsize}
\end{subfigure}%
\begin{subfigure}{.25\textwidth}
  \centering
  \includegraphics[width=1.0\linewidth]{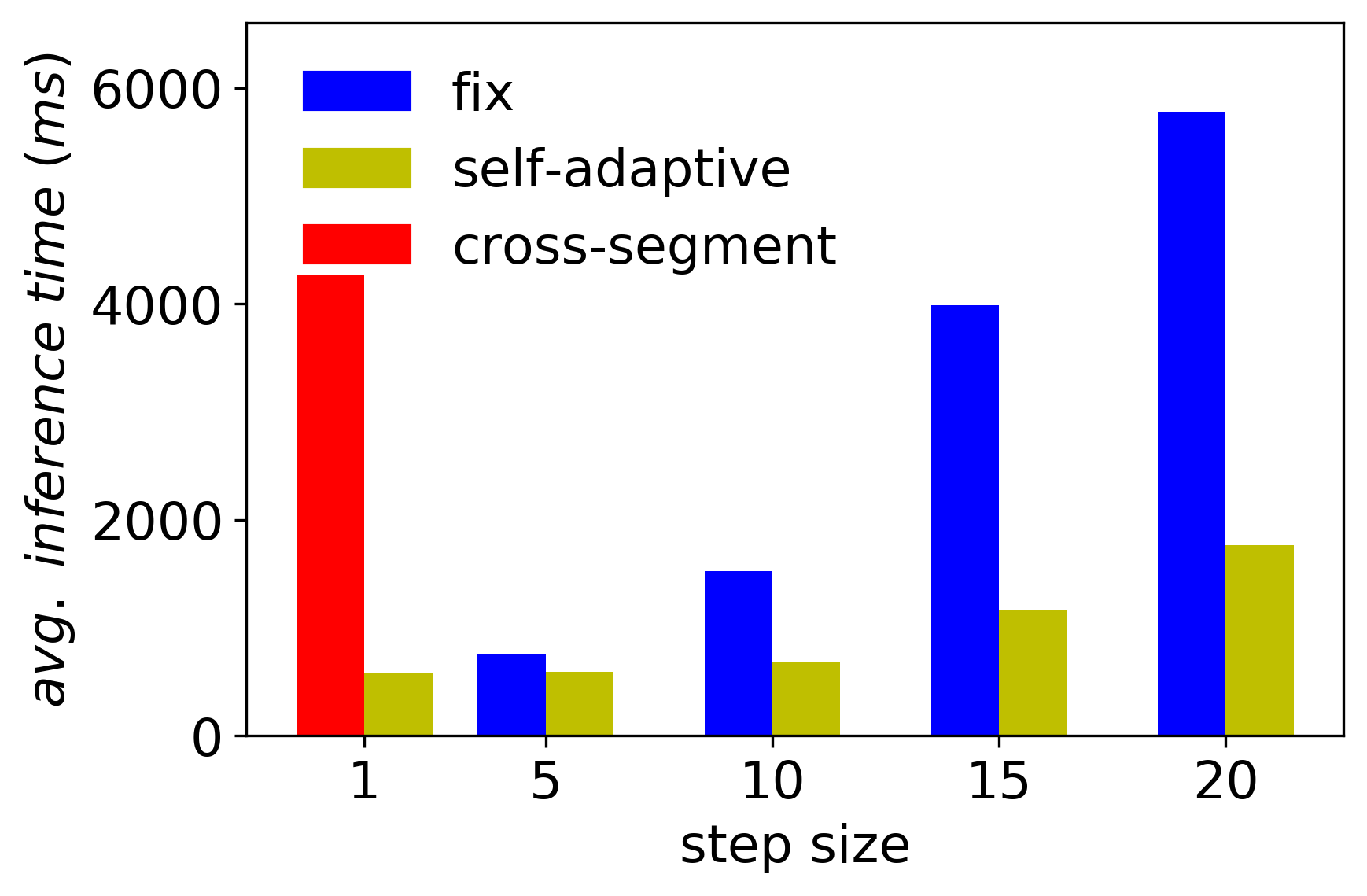}
  \caption{Inference time VS step size.}
  \label{fig:inference_windowsize}
\end{subfigure}
\caption{F$_1$ and avg. inference time (ms) per document for SeqModel with fixed and self-adaptive sliding window w.r.t. difference max backward step sizes (\textbf{step size}) and comparison to cross-segment BERT-Base 128-128.}
\label{fig:windowsize}
\end{figure}

\textbf{Effect of adding phone embeddings.} 
The effect of adding phone embeddings (denoted by +phone) on spoken document segmentation accuracy is shown in Table~\ref{tab:phoneEmb}. On the two spoken document datasets SD-zh-SP and SD-zh-MP, SeqModel:BERT-Base+phone notably outperforms SeqModel:BERT-Base by \textbf{2.8} and \textbf{2.1} points gains on F$_1$, both gains being statistically significant. For SeqMode:StructBERT-Base, +phone does not obtain improvement on SD-zh-SP but obtains +0.8 F$_1$ gain on SD-zh-MP. 
We observe that after fine-tuning with +phone, the input embeddings of words with similar pronunciations become close. Hence, this approach significantly improves robustness of the input embedding layer to ASR errors, and in turn, robustness of SeqModel.

\textbf{Effect of self-adaptive sliding window.} 
We randomly sample 1K documents from the Wiki-727 test set and measure the average inference time (ms) per document\footnote{Machine for inference is Intel(R) Xeon(R) Platinum 8269CY CPU @ 2.50GHz.  Intra\_op\_parallelism\_threads and inter\_op\_parallelism\_threads are set to 8 and 2, respectively.}. Figure~\ref{fig:f1score_windowsize} and ~\ref{fig:inference_windowsize} show F$_1$ and inference time for SeqModel using different max backward step sizes (shortened to step sizes) for fixed and self-adaptive sliding window during inference. For fixed sliding window, the next window starts from the last sentence in the previous window - max backward step size + 1, to be comparable to self-adaptive sliding window. For both sliding windows for SeqModel, the window size is 512 tokens. Cross-segment BERT-Base (with contexts 128-128) can be considered as using step size 1 and window size 256, and its inference time is 4169 ms/doc. SeqModel with fixed sliding window-step size 5 achieves 762 ms/doc, 81.7\% relative inference speedup. As analyzed in Section~\ref{subsec:inference}, self-adaptive sliding window dynamically discards irrelevant history for segmentation. When the step size increases, fixed sliding window includes more irrelevant content, hence we observe significant increase in inference time and drastic F$_1$ degradation (step size beyond 5).  In contrast, for self-adaptive sliding window, when the step size increases, the increase in inference time is much smaller and slower, and F$_1$ improves slightly yet steadily. These results demonstrate the advantages of self-adaptive sliding window on inference efficiency and accuracy.  Replacing fixed sliding window-step size 5 by self-adaptive sliding window-step size 10 further reduces SeqModel inference time to 689 ms/doc, less than \textbf{1/6} of inference time of cross-segment BERT-Base and a further 9.6\% relative speedup, while retaining the same F$_1$. 

\textbf{Readability Evaluations.} 
Spoken document segmentation aims at improving readability of ASR transcripts. Inspired by~\cite{DBLP:conf/interspeech/JonesWGWFRZ03,DBLP:conf/naacl/Shugrina10}, we design a reading comprehension evaluation since reading comprehension efficiency and accuracy strongly correlate with readability. We randomly sample 8 documents from SD-zh-SP and design 5-8 (on average 6) single-choice questions for each document based on its content. Every document $D_i$ is segmented by each segmentation model $M_j$.  The segmented document $D_i^j$ by applying $M_j$ on $D_i$ creates 50 crowd-sourcing jobs, in total $8 \times 50=400$ jobs for $M_j$, and each worker only works on documents segmented by one model. For each $M_j$, we select effective returns as returns with all questions for one document answered and the recorded duration $d$ for answering all questions satisfying $6$ minutes $\leq d \leq 20$ minutes. 
We obtain on average 87 effective returns among the 400 jobs for each $M_j$, probably due to the challenges posed by long-form spoken documents and the complex finance domain. The average number of answers to compute answer accuracy for each model is $87  \times 6  = 522$ samples. SeqModel:BERT-Base demonstrates better readability by achieving 64.5\% answer accuracy, compared to 59.5\% from cross-segment BERT-Base.

\vspace{-2mm}
\section{Conclusions}
We propose a sequence model with self-adaptive sliding window for document segmentation. Our model significantly outperforms current SOTAs on an English benchmark and Chinese written text and spoken document datasets both on accuracy and inference latency.  We propose a phone-embedding based approach and significantly improve spoken document performance. Future work includes studying more effective structural modeling for long-form spoken documents.

\bibliographystyle{IEEEbib}
\bibliography{mybib}

\end{document}